\documentclass{article}

\usepackage{amsmath}
\usepackage{amssymb}
\usepackage{graphicx}
\usepackage{hyperref}
\usepackage{multirow}
\usepackage{paralist}
\usepackage{spconf}
\usepackage{subcaption}
\usepackage{tikz}
\usetikzlibrary{automata, positioning, arrows}
\tikzset{
    ->, 
    >=stealth', 
    node distance=2cm, 
    initial text=$ $, 
}

\title{LAST: Scalable Lattice-Based Speech Modelling in JAX}
\name{Ke Wu, Ehsan Variani, Tom Bagby, Michael Riley}
\address{Google Research \\ \{wuke, variani, tombagby, riley\}@google.com}
\begin{document}
%
\maketitle
\begin{abstract}
We introduce LAST, a LAttice-based Speech Transducer library in JAX.
With an emphasis on flexibility, ease-of-use, and scalability, LAST implements differentiable weighted finite state automaton (WFSA) algorithms needed for training \& inference that scale to a large WFSA such as a recognition lattice over the entire utterance.
Despite these WFSA algorithms being well-known in the literature, new challenges arise from performance characteristics of modern architectures, and from nuances in automatic differentiation.
We describe a suite of generally applicable techniques employed in LAST to address these challenges, and demonstrate their effectiveness with benchmarks on TPUv3 and V100 GPU.
\end{abstract}


\section{Introduction}
\label{sec:introduction}

Weighted finite state automata (WFSA) have been a vital tool in building automatic speech recognition (ASR) systems for both model training \& inference.
Common neural ASR models (CTC~\cite{graves2006connectionist}, RNN-T~\cite{graves2012sequence}) exhibit a close connection to the WFSA formalism, and there has been a renewed interest in casting these models into WFSAs, e.g. frameworks such as k2~\cite{povey2021speech}, GTN~\cite{hannun2020differentiable}, and PyTorch-Struct~\cite{rush2020torch}.
More recently, \cite{variani2022global} demonstrates globally normalized sequence models built within the WFSA formalism can greatly reduce the word error rate of streaming ASR models.

Motivated by the need for efficient yet flexible and easy-to-use sequence modelling tools for implementing the aforementioned methods, especially the globally normalized models such as \cite{variani2022global}, we developed LAST, a lattice-based speech transducer library in JAX~\cite{jax2018github}.
LAST provides a simple interface for operations over a recognition lattice, and is fully compatible with core features in JAX such as automatic differentiation and function transformations.
Being written in pure JAX allows LAST to run efficiently on CPUs, GPUs, and TPUs without any code modification, whereas the existing WFSA based sequence modelling toolkits \cite{hannun2020differentiable,povey2021speech} employ handwritten kernels targeting a limited set of hardware.

In this paper, we introduce the design of LAST, discuss the technical challenges that drive crucial design choices, and present benchmark results.
LAST is open source software under the Apache License 2.0.
The complete code can be found at {\small \url{https://github.com/google-research/last}}.

\section{Background}
\label{sec:background}

\subsection{Finite State Automata}

We refer readers to \cite{mohri2008, mohri2009} for a comprehensive introduction to finite state automata and related algorithms.
A weighted finite state automaton (WFSA) $A = (\Sigma, Q, i, f, E)$ over a semiring $(\mathbb{K}, \oplus, \otimes, \bar{0}, \bar{1})$ is specified by
\begin{inparaitem}[]
    \item a finite alphabet $\Sigma$,
    \item a finite set of states $Q$,
    \item an initial state $i \in Q$,
    \item a final state $f$,
    \item and a finite set of arcs $E \subseteq Q \times (\Sigma \cup \{\epsilon\}) \times \mathbb{K} \times Q$ ($\epsilon$ denotes the empty label sequence).
\end{inparaitem}
We denote
\begin{inparaitem}[]
    \item the set of accepting paths $\Pi$;
    \item an arc $e$'s weight $w[e]$;
    \item $\mathrm{next[e]}$ and $\mathrm{prev[e]}$ the source and destination states of arc $e$. 
\end{inparaitem}
WFSA are also commonly refered to as \emph{lattices} in the ASR literature.
The most commonly used WFSA algorithms in building ASR systems are
\begin{compactdesc}
    \item[Intersection] A WFSA is a compact representation of a set of weighted strings from $\Sigma^* \times \mathbb{K}$. The intersection of two weighted sets representable as WFSAs is equivalent to the intersection of the corresponding WFSAs.
    \item[Shortest distance] The shortest distance of a WFSA is the $\oplus$-sum of the weights of all paths from state $i$ to state $f$ (\emph{accepting paths}).
    Most ASR lattices are acyclic WFSAs by construction, and their shortest distance can be computed by visiting the states in topological order.
    \item[Shortest path] The shortest path of a WFSA is the accepting path with lowest cost or highest score.
    It can be computed for acyclic WFSAs in a similar fashion as shortest distance.
\end{compactdesc}

For example, the computation of the RNN-T loss \cite{graves2012sequence} can be broken down into two WFSA operations:
\begin{inparaenum}[(1)]
    \item Intersecting the (infinite) complete recognition lattice with the reference output string (Figure 1 of \cite{graves2012sequence});
    \item Shortest distance on the intersection.
\end{inparaenum}
Another example is the computation of the globally normalized loss in \cite{variani2022global}:
\begin{inparaenum}[(1)]
    \item Shortest distance on the complete recognition lattice to obtain the denominator;
    \item Similar to RNN-T, intersection with the reference and then shortest distance to obtain the numerator.
\end{inparaenum}

During training, we differentiate through shortest distance \& intersection to obtain the gradients of the training loss with respect to the arc weights.

\subsection{JAX and Automatic Differentiation}

We chose to implement LAST in JAX because of the performance, ease of use, and flexibility JAX offers.
JAX \cite{jax2018github} is a high-performance machine learning research toolkit that has seen a growing popularity.
JAX programs written in the NumPy API execute efficiently on CPUs, GPUs and TPUs.
Similar to \cite{paszke2019pytorch}, the vector-Jacobian-product function needed for (reverse mode) automatic differentiation \cite{linnainmaa1976taylor,griewank2008evaluating} can be either automatically derived for any function expressed in differentiable JAX primitives, or manually defined.

\subsection{Gradients of Shortest Distance}
\label{sec:gradients-of-shortest-distance}

With the WFSA topology fixed, the shortest distance $D = \bigoplus_{\pi \in \Pi} \left( \bigotimes_{e \in \pi} w[e] \right)$ is a function of arc weights $w[e], \forall e \in E$.
Under the log semiring, $\frac{\partial D}{\partial w[e]}$ is the marginal probability of the arc $e$, viewing path weights as unnormalized log-probabilities,
\begin{equation}
    \label{eq:arc-marginal}
    \frac{\partial D}{\partial w[e]} = \sum_{\pi \in \Pi \land e \in \pi} \frac{\exp\left(\sum_{e' \in \pi} w[e']\right)}{\exp(D)}.
\end{equation}
The forward-backward algorithm \cite{baum1972inequality} is commonly used to compute this quantity for acyclic WFSAs, while automatic differentiation can also be applied to achieve the same goal in the same time complexity \cite{eisner2016inside} (more on this in Section~\ref{sec:auto-diff-vs-forward-backward}).

Moreover, under the tropical semiring, $\frac{\partial D}{\partial w[e]}$ is a 0/1 mask denoting if $e$ is on the shortest path.
A differentiable shortest distance algorithm thus greatly simplifies the implementation of inference: we get the shortest path algorithm for free.
This technique of deriving shortest path by differentiating shortest distance is also adopted by k2 \cite{povey2021speech} and PyTorch-Struct \cite{rush2020torch}.

\section{Library Design}
\label{sec:library-design}

The GNAT model outlined in \cite{variani2022global} is a modular theoretical framework encompassing all the common neural speech recognition models.
\cite{variani2022global} reports that the globally normalized GNAT model greatly reduces the word error rate of streaming ASR.
The primary goal of LAST is to replicate GNAT.
Therefore, the top level API of LAST closely follows components of the GNAT model.
GNAT breaks down modelling choices into 3 components:
\begin{inparadesc}
    \item[The context dependency] is an unweighted deterministic finite automaton that accepts $\Sigma^*$.
    The states in a context dependency encode the output history.
    \item[The alignment lattice] for a given input length $T$ is an acyclic unweighted finite automaton.
    The states in an alignment lattice encode the correspondence between input frames and output alignment labels.
    \item[The weight function] is a function $Q_T \times Q_C \times (\Sigma \cup \{\epsilon\}) \rightarrow \mathbb{K}$.
    The intersection of the context dependency and the alignment lattice defines the topology of the recognition lattice, whereas the weight function defines the weights of each arc.
\end{inparadesc}
Figures~\ref{fig:n-gram-context},~\ref{fig:frame-dependent}, and~\ref{fig:recognition-lattice} are a concrete example of these components from \cite{variani2022global}.

\begin{figure}
    \centering
    \begin{subfigure}[t]{0.4\linewidth}
        \centering
        \scalebox{0.5}{%
        \begin{tikzpicture}
            \node[state, initial, accepting] (empty) {$\epsilon$};
            \node[state, accepting, below left of=empty, yshift=-1cm] (a) {$a$};
            \node[state, accepting, below right of=empty, yshift=-1cm] (b) {$b$};
    
            \draw (empty) edge[bend right, above left] node{$a$} (a)
                  (empty) edge[bend left, above right] node{$b$} (b)
                  (a) edge[loop left] node{$a$} (a)
                  (a) edge[bend left, above] node{$b$} (b)
                  (b) edge[bend left, above] node{$a$} (a)
                  (b) edge[loop right] node{$b$} (b);
        \end{tikzpicture}
        }%
        \caption{Context size 1 $n$-gram context dependency with $\Sigma=\{a,b\}$.}
        \label{fig:n-gram-context}
    \end{subfigure}
    \hfill
    \begin{subfigure}[t]{0.5\linewidth}
        \centering
        \scalebox{0.5}{%
        \begin{tikzpicture}
            \node[state, initial] (q0) {0};
            \node[state, right of=q0] (q1) {1};
            \node[state, right of=q1] (q2) {2};
            \node[state, accepting, right of=q2] (q3) {3};
    
            \draw (q0) edge[bend right, above] node{$\Sigma$} (q1)
                  (q0) edge[bend left, above] node{$\epsilon$} (q1)
                  (q1) edge[bend right, above] node{$\Sigma$} (q2)
                  (q1) edge[bend left, above] node{$\epsilon$} (q2)
                  (q2) edge[bend right, above] node{$\Sigma$} (q3)
                  (q2) edge[bend left, above] node{$\epsilon$} (q3);
        \end{tikzpicture}
        }
        \caption{Frame dependent alignment lattice with $T=3$.}
        \label{fig:frame-dependent}
    \end{subfigure}
    \begin{subfigure}{0.9\linewidth}
        \centering
        \scalebox{0.5}{%
        \begin{tikzpicture}
            \tikzstyle{highlight}=[line width=1.5pt];
    
            \node[state, initial] (e0) {$0,\epsilon$};
            \node[state, right of=e0] (a1) {$1,a$};
            \node[state, above of=a1] (e1) {$1,\epsilon$};
            \node[state, below of=a1] (b1) {$1,b$};
            \node[state, right=3cm of e1] (e2) {$2,\epsilon$};
            \node[state, below of=e2] (a2) {$2,a$};
            \node[state, below of=a2] (b2) {$2,b$};
            \node[state, accepting, right=3cm of e2] (e3) {$3,\epsilon$};
            \node[state, accepting, below of=e3] (a3) {$3,a$};
            \node[state, accepting, below of=a3] (b3) {$3,b$};
    
            \draw (e0) edge[bend left, above, highlight] node{$\boldsymbol{\epsilon}$} (e1)
                  (e0) edge[above, highlight] node{$\boldsymbol{a}$} (a1)
                  (e0) edge[bend right, above] node{$b$} (b1)
                  (e1) edge[bend left, above] node{$\epsilon$} (e2)
                  (e1) edge[bend left, above, highlight] node{$\boldsymbol{a}$} (a2)
                  (e1) edge[bend left, above] node{$b$} (b2)
                  (a1) edge[below, highlight] node{$\boldsymbol{\epsilon}$} (e2)
                  (a1) edge[above] node{$a$} (a2)
                  (a1) edge[above, highlight] node{$\boldsymbol{b}$} (b2)
                  (b1) edge[bend right, above] node{$\epsilon$} (e2)
                  (b1) edge[bend right, above] node{$a$} (a2)
                  (b1) edge[bend right, above] node{$b$} (b2)
                  (e2) edge[bend left, above] node{$\epsilon$} (e3)
                  (e2) edge[bend left, above] node{$a$} (a3)
                  (e2) edge[bend left, above, highlight] node{$\boldsymbol{b}$} (b3)
                  (a2) edge[below] node{$\epsilon$} (e3)
                  (a2) edge[above] node{$a$} (a3)
                  (a2) edge[above, highlight] node{$\boldsymbol{b}$} (b3)
                  (b2) edge[bend right, above, highlight] node{$\boldsymbol{\epsilon}$} (e3)
                  (b2) edge[bend right, above] node{$a$} (a3)
                  (b2) edge[bend right, above] node{$b$} (b3)
                  ;
        \end{tikzpicture}
        }
        \caption{Recognition lattice: intersection of Figure~\ref{fig:n-gram-context} and Figure~\ref{fig:frame-dependent} with all paths for output $ab$ highlighted. The weights (omitted here) are defined by passing the states and labels to the weight function.}
        \label{fig:recognition-lattice}
    \end{subfigure}
\end{figure}

The core of the LAST library, class \textsc{RecognitionLattice} is a \texttt{flax.linen} module \cite{flax} that combines these 3 components to produce the actual states and weights for any recognition lattice induced from a sequence of input frames.
The shortest distance of either the entire \textsc{RecognitionLattice}, or its intersection with an output string, can then be obtained through API methods for training locally or globally normalized models.
A low level state/arc-based API is also provided for applications such as beam search decoding.

While class \textsc{RecognitionLattice} provides efficient high level operations through a simple interface by hiding implementation details, the modelling components are instead designed as easy to extend interfaces,
\begin{compactitem}
    \item In addition to the \textsc{FullNGram} context dependency described in \cite{variani2022global}, there is also a \textsc{NextStateTable} context dependency, enabling the use of any context dependency topology with a look-up table.
    \item Both \textsc{FrameDependent} and \textsc{FrameLabelDependent} alignment lattices are available.
    \item Weight functions can be implemented as the combination of a \textsc{WeightFnCacher} that performs precomputation of static data (such as context state embeddings), and a \textsc{WeightFn} that actually produces arc weights.
\end{compactitem}

WFSAs in LAST are not just defined with data (i.e. arrays of state/arc tables) but also with code that describes higher level operations.
This is a key difference of LAST from prior art such as k2 \cite{povey2021speech}. 
For example, \textsc{ContextDependency} provides \textsc{ForwardReduce} and \textsc{BackwardBroadcast}, simple yet crucial primitives that \textsc{RecognitionLattice} uses to produce a highly vectorized forward-backward implementation (Section~\ref{sec:random-access}).
Such higher level APIs are designed to take advantage of special structures in certain WFSAs to achieve higher efficiency with minimum interface complexity.
    
\section{Challenges}
\label{sec:challenges}

\begin{table}
    \centering
    \begin{tabular}{p{8em}|c|c}
        Lattice & $|Q|$ & $|E|$ \\
        \hline
        RNN-T, intersected & $O(T U)$ & $O(T U)$ \\
        \hline
        GNAT, intersected & $O(T U)$ & $O(T U)$ \\
        \hline
        GNAT, complete & $O(T C)$ & $O(T C V)$        
    \end{tabular}
    \caption{Lattice sizes. $T$: input length, $U$: output length, $C$: number of context dependency states, $V$: vocabulary size.}
    \label{tab:lattice-sizes}
\end{table}

The main challenge in scaling up differentiable WFSA algorithms for modern accelerator architectures is memory: the memory is more limited in size compared to CPU DRAM, and memory access can be much slower than compute.
Efficient memory use is thus behind many design details in LAST.

\subsection{The Need for On-the-Fly Arc Weight Computation}

Existing frameworks such as k2 \cite{povey2021speech} or GTN \cite{hannun2020differentiable} store all the arc weights of a WFSA in memory at once.
Table~\ref{tab:lattice-sizes} summarizes the number of states $|Q|$ and the number arcs $|E|$ of various types of lattices.
When the lattice is obtained from an intersection with a reference output, $|Q|$ and $|E|$ are comparable.
Storing all the arc weights at once in memory is thus practical for training locally normalized models which only involves the lattice after intersection.
However, $|E|$ can be much larger than $|Q|$ for a complete recognition lattice needed in training globally normalized models, and storing all $O(TCV)$ arc weights quickly become impractical.

\textbf{Remedy} The highly structured nature of ASR lattices is very suitable for on-the-fly arc weight computation.
The states are $(t, q)$ pairs of time $t$ and additional state $q$.
The arc weights can be computed in the increasing order of $t$ during the topological visit in shortest distance.

\subsection{Standard Automatic Differentiation Is Memory Inefficient under the Log Semiring}
\label{sec:auto-diff-vs-forward-backward}

\cite{eisner2016inside} points out that automatic differentiation over shortest distance gives rise to the classical forward-backward algorithm \cite{baum1972inequality}.
However \cite{eisner2016inside} only analyzed the algorithms under the real semiring.
We shall show that under the log semiring, naive automatic differentiation yields an algorithm that requires $O(|E| + |Q|)$ space, whereas the forward-backward algorithm only requires $O(|Q|)$.
Notably, this space complexity holds when the arc weights are either a part of the input (not included in space complexity), or computed on-the-fly.

For an acyclic WFSA, the following forward weight recurrence computed for each $q \in Q$ in a topological order gives the shortest distance $D = \alpha[f]$,
\begin{equation}
    \label{eq:forward-recurrence}
    \alpha[q] = \bigoplus_{\mathrm{next}[e] = q} \left(\alpha[\mathrm{prev}[e]] \otimes w[e]\right).
\end{equation}

Automatic differentiation, a.k.a. back propagation \cite{rumelhart1986learning}, computes the partial derivative $\frac{\partial D}{\partial w[e]}$
by repeatedly computing $\frac{\partial \alpha[\mathrm{next}[e]]}{\partial w[e]}$ and $\frac{\partial \alpha[\mathrm{next}[e]]}{\partial \alpha[\mathrm{prev}[e]]}$ in a reverse topological order and then applying the chain rule $\frac{\partial D}{\partial w[e]} = \sum_{q \in Q} \frac{\partial D}{\partial \alpha[q]} \frac{\partial \alpha[q]}{\partial w[e]}$.

Computing $\frac{\partial \alpha[\mathrm{next}[e]]}{\partial w[e]}$ requires knowing the weights of arcs leading to $\mathrm{next}[e]$ and the forward weights of the source states of these arcs.
Because $w[e]$ is either a part of the input, or computed on-the-fly, only $\alpha[q]$ is stored, taking space $O(|Q|)$.
The situation is different for $\frac{\partial \alpha[\mathrm{next}[e]]}{\partial \alpha[\mathrm{prev}[e]]}$, which we shall analyze separately for the real, log, and tropical semirings.
For simplicity, let $\Gamma[q]$ be the set of states with arcs leading to state $q$, we denote the partial derivative term $\frac{\partial \alpha[\mathrm{next}[e]]}{\partial \alpha[\mathrm{prev}[e]]}$ alternatively as $\frac{\partial \alpha[q]}{\partial \alpha[p]}$ for $p \in \Gamma[q]$ (for states not in $\Gamma[q]$, the partial derivative is not used in automatic differentiation).

\subsubsection{Real semiring}

In this case, $\alpha[q] = \sum_{\mathrm{next}[e] = q} \alpha[\mathrm{prev}[e]] w[e]$, thus
$
    \frac{\partial \alpha[q]}{\partial \alpha[p]} = \sum_{\mathrm{prev[e]}=p \land \mathrm{next[e]}=q}w[e].
$
Since we just need the arc weights to compute the relevant gradients, the overall space complexity remains $O(|Q|)$.

\subsubsection{Log semiring}

In this case, $\alpha[q] = \log\left(\sum_{\mathrm{next}[e] = q} \exp(\alpha[\mathrm{prev}[e]] + w[e])\right)$, thus
$
    \frac{\partial \alpha[q]}{\partial \alpha[p]} = \sum_{\mathrm{prev[e]}=p \land \mathrm{next[e]}=q} \frac{\exp(\alpha[p] + w[e])}{\exp\alpha[q]}.
$
Forward-backward utilizes the following fact to eliminate the need of storing $\alpha[p] + w[e]$,
\begin{align*}
    & \frac{\partial \alpha[q]}{\partial \alpha[p]}\frac{\partial \alpha[p]}{\partial \alpha[r]} \\
    = & \left( \sum_{\mathrm{prev[e_2]}=p \land \mathrm{next[e_2]}=q} \frac{\exp(\alpha[p] + w[e_2])}{\exp\alpha[q]} \right) \times \\
    & \left( \sum_{\mathrm{prev[e_1]}=r \land \mathrm{next[e_1]}=p} \frac{ \exp(\alpha[r] + w[e_1])}{\exp\alpha[p]} \right) \\
    = & \sum_{\substack{\mathrm{prev[e_2]}=p \land \mathrm{next[e_2]}=q \\ \mathrm{prev[e_1]}=r \land \mathrm{next[e_1]}=p}} \frac{\exp(w[e_2] + w[e_1] + \alpha[r])}{\exp\alpha[q]}.
\end{align*}
Repeatedly chaining the partial derivatives from the final state $f$ to  $\mathrm{prev}[e]$ results in a summation over all the paths from $\mathrm{prev}[e]$ to $f$, giving rise to Equation~\ref{eq:arc-marginal}.
To compute Equation~\ref{eq:arc-marginal}, the forward-backward algorithm stores the backward weights $\beta[q]$, the sum of path weights from all states to the final state, in addition to $\alpha[q]$, with the space complexity remaining $O(|Q|)$.
In contrast, an automatic differentiation engine can not perform the symbolic optimization that cancels out $\exp\alpha[p]$ from the equation due to its mechanical nature.
It thus need to store all the intermediate $\alpha[p] + w[e]$ values from the forward pass to use in the backward pass, leading to an overall $O(|E| + |Q|)$ space complexity, regardless of whether the arc weights are part of the input, or computed on-the-fly.

\subsubsection{Tropical semiring}

In this case, $\alpha[q] = \max_{\mathrm{next}[e] = q} (\alpha[\mathrm{prev}[e]] + w[e])$, thus (the sub-gradient)
$
    \frac{\partial \alpha[q]}{\partial \alpha[p]} = \mathbf{1}_{\alpha[p] + w[e] = \alpha[q]}.
$
The vector-Jacobian-product of $\max$ is implemented by storing the index $\arg\max_p \alpha[p] + w[e]$ for all $q$, taking up just $O(|Q|)$ space.

\subsubsection{Remedy: Gradient Rematerialization}
\label{sec:remat}

JAX offers a flexible gradient rematerialization (also called checkpointing \cite{bennett1973logical,griewank2008evaluating}) mechanism to control what intermediate values can be saved during the forward pass.
Whatever not saved is recomputed during the backward pass, thereby trading memory usage with compute.
We can thus disallow any saving of intermediate values within individual invocations of Equation~\ref{eq:forward-recurrence} simply through a \texttt{jax.remat} call.
In Section~\ref{sec:benchmarks}, we compare gradient rematerialization with the hand-written forward-backward algorithm.

\section{Benchmarks}
\label{sec:benchmarks}

\begin{table}
    \centering
    \begin{tabular}{l|p{12em}}
        Vocabulary size & 32 (i.e. grapheme) \\
        \hline
        Context dependency & \textsc{FullNGram}, context size 2, 1057 states \\
        \hline
        Alignment lattice & \textsc{FrameDependent} \\
        \hline
        Weight function & \textsc{SharedEmb}, 512 hidden units \\
        \hline
        Batch size & 16 \\
        \hline
        Input length & 1024 \\
        \hline
        Output length & 256 \\
    \end{tabular}
    \caption{Model configurations and input size}
    \label{tab:model-data-configs}
\end{table}

\begin{table}
    \centering
    \begin{tabular}{l|p{6em}|p{4em}|p{4em}}
        \multicolumn{2}{l|}{Acclerator} & TPUv3 & V100 \\
        \hline
        Training & Forward-backward & 0.92s / 217MB & 2.42s / 278MB \\
        \cline{2-4}
        & Auto-diff with remat & 1.12s / 219MB & 2.40s / 278MB \\
        \cline{2-4}
        & Auto-diff without remat & OOM / $\geq$ 102G & OOM / $\geq$ 68.5G \\
        \hline
        \multicolumn{2}{l|}{Inference} & 0.41s / 171MB & 0.30s / 203MB \\
        \hline
        \multicolumn{2}{l|}{k2 Training} & & 1.69s / 11.9G \\
        \hline
        \multicolumn{2}{l|}{k2 Inference} & & 0.70s / 9.6G \\
    \end{tabular}
    \caption{Benchmark results (speed and memory usage)}
    \label{tab:training-speed}
\end{table}

We benchmark the training and inference speed and memory usage on a single Google TPUv3 core and a single nVidia V100 GPU.
Table~\ref{tab:model-data-configs} outlines the configurations of the globally normalized GNAT model and benchmark data.
The setup amounts to approximately 1 million parameters in the ``decoder'' component of an encoder-decoder architecure.
Each input in the batch induces a lattice of about 1 million states and 33 million arcs.
As a comparison, we implement the same model in k2 \cite{povey2021speech}.
Even with optimizations such as \textsc{DenseFsaVec} and gradient rematerialization, our k2 implementation is not able to process batch size 16 input without running out of memory.
We thus split the input into smaller batches (5+5+6) and accumulate gradients during training. 

Table~\ref{tab:training-speed} lists the benchmark results.
LAST only implements the ``decoder'' component in an encoder-decoder architecture, we thus benchmark the gradient computation from the loss value to the decoder parameters and the encoder frames, using various strategies for computing the shortest distance gradients as discussed in Section~\ref{sec:remat}.
Consistent with our analysis in Section~\ref{sec:auto-diff-vs-forward-backward}, the forward-backward algorithm uses far less memory than automatic differentiation without gradient rematerialization.
Because we ran out of memory on both TPUv3 and V100 when trying to run automatic differentiation without gradient rematerialization, we report the amount of projected memory usage reported by the XLA compiler when it aborted the compilations.
Forward-backward and automatic differentiation with rematerialization use a similar amount of memory.
On TPUv3, forward-backward is 18\% faster than automatic differentiation with gradient rematerialization, whereas on GPU, there is no significant speed difference.
For inference speed, we benchmark shortest path derived from differentiating shortest distance.
Despite a drastic 2.4x difference in training speed, the V100 GPU is faster at inference.
These differences in the speed are likely due to different hardware design trade-offs.

The memory usage of our implementation in k2 shows that on-the-fly arc weight computation can bring about significant memory savings.
k2's training speed is faster than LAST because computing all arc weights at once affords more opportunity for parallelization: we compute arc weights 1 frame at a time in LAST, but 16 frames at a time in k2.
Had we also computed arc weights for 1 or 2 frames at a time in k2, the training time would increase to 4.76s or 2.59s respectively.

\section{Conclusion}

We presented the LAST library and discussed challenges in scaling up WFSA algorithms in an automatic differentiation framework.
These challenges led us to design LAST differently from prior art.
Our generally applicable techniques to counter these challenges proved effective as shown in the benchmark results.

\bibliographystyle{IEEEbib}
\bibliography{last}

\end{document}